\documentclass{article}

\usepackage[final]{corl_2019} 

\usepackage{enumerate}
\usepackage{amssymb}
\usepackage{amsmath}
\usepackage{amsfonts}
\usepackage{graphicx}
\usepackage{color}
\usepackage[subrefformat=parens]{subcaption}
\usepackage{subcaption}
\usepackage{caption}
\usepackage{multirow}
\usepackage{multicol}
\usepackage{tikz}
\usetikzlibrary{calc}
\usepackage{subfloat}
\usepackage{verbatim}
\usepackage[ruled,linesnumbered]{algorithm2e}
\usepackage{hyperref} 

\newcommand{\Stateset}{\mathcal{S}}
\newcommand{\Actionset}{\mathcal{A}}
\newcommand{\transitionfunc}{P}
\newcommand{\policy}{\pi}
\newcommand{\newpolicy}{\tilde{\pi}}
\newcommand{\discountfactor}{\gamma}
\newcommand{\statet}{s_t}
\newcommand{\statetplusone}{s_{t+1}}
\newcommand{\reward}{r}
\newcommand{\actiont}{a_t}
\newcommand{\actiontplusone}{a_{t+1}}
\newcommand{\statezero}{s_0}
\newcommand{\actionzero}{a_0}
\newcommand{\state}{s}
\newcommand{\action}{a}
\newcommand{\advantage}{A}
\newcommand{\advantagepi}{\advantage_{\pi}}

\newcommand{\expectedreturn}{\eta}
\newcommand{\expectation}{\mathbb{E}}
\newcommand{\statedistribution}{\rho}
\newcommand{\qfunctionpi}{Q_{\pi}}
\newcommand{\valuefunctionpi}{V_{\pi}}
\newcommand{\optparameter}{\theta}
\newcommand{\optparameterold}{\theta_{\text{old}}}

\newcommand{\dklestimate}{\bar{D}_{KL}}

\newcommand{\half}{\frac{1}{2}}
\newcommand{\urho}{\underline{\nu}}
\newcommand{\orho}{\overline{\nu}}
\newcommand{\uomega}{\underline{\omega}}
\newcommand{\oomega}{\overline{\omega}}
\newcommand{\odelta}{\overline{\delta}}
\newcommand{\ukappa}{\underline{\kappa}}

\newcommand{\qntrm}{QNTRM}
\newcommand{\qntrpo}{QNTRPO}

\title{Quasi-Newton Trust Region Policy Optimization}

\author{
  Devesh K. Jha\\
  MERL\\
  Cambridge, MA\\
  \texttt{jha@merl.com} \\
  \And
  Arvind U. Raghunathan\\
  MERL\\
  Cambridge, MA\\
  \texttt{raghunathan@merl.com} \\
  \And
  Diego Romeres \\
  MERL\\
  Cambridge, MA\\
  \texttt{romeres@merl.com} \\
}

\begin{document}
\maketitle


\begin{abstract}
We propose a trust region method for policy optimization that employs Quasi-Newton approximation for the Hessian, called Quasi-Newton Trust Region Policy Optimization (\qntrpo). Gradient descent is the de facto algorithm for reinforcement learning tasks  with continuous controls. The algorithm has achieved state-of-the-art performance 
when used in reinforcement learning across a wide range of tasks. However, the algorithm suffers from a number of drawbacks including: lack of stepsize selection criterion, and slow convergence.  We investigate the use of a trust region method using dogleg step and a Quasi-Newton approximation for the Hessian for policy optimization. We demonstrate through numerical experiments over a wide range of challenging continuous control tasks that our particular choice is efficient in terms of number of samples and improves performance\footnote{Codes for the proposed method could be downloaded from \url{http://www.merl.com/research/?research=license-request&sw=QNTRPO}}.  
\end{abstract}

\keywords{Reinforcement Learning, Trust Region Optimization, Quasi-Newton Methods, Policy Gradient} 

\section{Introduction}\label{sec:introduction}
Reinforcement Learning (RL) is a learning framework that handles sequential decision-making problems, wherein an `agent' or decision maker learns a policy to optimize a long-term reward by interacting with the (unknown) environment. At each step, an RL agent obtains evaluative feedback (called reward or cost) about the performance of its action, allowing it to improve (maximize or minimize) the performance of subsequent actions \cite{sutton1998reinforcement}. Recent research has resulted in remarkable success of these algorithms in various domains like computer games~\cite{silver2016mastering,silver2017mastering}, robotics~\cite{levine2016end,duan2016benchmarking}, etc. 
Policy gradient algorithms can directly optimize the cumulative reward and can be used with a lot of different non-linear function approximators including neural networks. 
Consequently, policy gradient algorithms are appealing 
for a lot of different applications, and are widely used for a lot of robotic applications~\cite{peters2006policy, kober2009policy, deisenroth2013survey}. 
As a result, it has attracted significant attention in the research community where several new algorithms have been proposed to solve the related problems. 
However, several problems remain open including monotonic improvement in performance of the policy, selecting the right learning rate (or step-size) during optimization, etc.  

Notably, the Trust Region Policy Optimization (TRPO) has been proposed to provide monotonic improvement of policy performance~\cite{schulman2015trust}. 
TRPO relies on a linear model of the objective function and quadratic model of the constraints to determine a candidate search direction.  Even though a theoretically justified trust region radius is derived such a radius cannot be computed and hence, linesearch is employed for obtaining a stepsize that ensures progress to a solution.  Consequently, TRPO is a scaled gradient descent algorithm and is not a trust region algorithm as the name suggests.  More importantly, TRPO does not inherit the flexibility and convergence guarantees provided by the trust region framework~\cite{NocedalWrightBook}.  As a consequence, the impact of trust region algorithms have not been fully investigated in the context of policy optimization. 

Our objective in this work is to show that a \emph{classical trust region method} in conjunction 
with \emph{quadratic model} of the objective addresses the drawbacks of TRPO.  
It is well known that incorporating curvature information of the objective function (i.e. quadratic approximation)  allows for rapid convergence in the neighborhood of a solution. Far from a solution, the curvature information should be incorporated in a manner that ensures the search direction improves on the reduction obtained by a linear model.  
We propose the \emph{Quasi-Newton Trust Region Policy Optimization} (\qntrpo) which uses a dogleg method for computing the step, i.e. both the search direction and stepsize are determined jointly. The Quasi-Newton (QN) method allows for incorporating curvature information by approximating the Hessian of the objective without the need for computing exact second derivatives.  In particular, we employ the \emph{classical BFGS} approximation~\cite{NocedalWrightBook}. The dogleg method is well known to produce at least as much reduction obtained using a linear model~\cite{NocedalWrightBook}, thus ensuring that \qntrpo\, does at least as well as the TRPO. The choice of QN method and search direction are chosen to ensure that global convergence properties are retained and the computational cost is comparable to that of TRPO.  
%
We want to investigate if \qntrpo, which has a different step from TRPO, can
\begin{enumerate}[(i)]
    \item accelerate the convergence to an optimal policy, and
    \item achieve better performance in terms of average reward.
\end{enumerate}

\qntrpo\, computes the stepsize as part of the search direction computation and stepsize is naturally varied according to the accuracy of the quadratic model of the objective. \qntrpo\, learns faster than TRPO due to the quadratic model and improved search direction. We test the proposed method on several challenging locomotion tasks for simulated robots in the OpenAI Gym environment. We compare the results against the original TRPO algorithm and show that we can consistently achieve better learning rate as well as performance.

\section{Background}\label{background}
We first introduce notation and summarize the standard policy gradient framework for RL and the TRPO problem.

\subsection{Notation}\label{subsec:notation}
We address policy learning in continuous/discrete action spaces. We consider an infinite horizon Markov decision process (MDP) defined by the tuple $(\Stateset,\Actionset,\transitionfunc,\reward,\discountfactor)$, where the state space $\Stateset$ is continuous, and the unknown state transition probability $\transitionfunc:\Stateset\times\Stateset\times\Actionset\rightarrow [0,1]$ represents the probability density of the next state $\statetplusone\in \Stateset$ given the current state $\statet \in \Stateset$ and action $\actiont \in \Actionset$ and $\discountfactor$ is the standard discount factor. The environment emits a reward $r: \mathcal{S}\times \Actionset\rightarrow \mathbb{R}$ on each transition. 

Let $\policy$ denote a stochastic policy $\policy: \Stateset \times \Actionset \rightarrow [0,1]$, and let $\expectedreturn (\policy)$ denote the expected discounted reward:

\begin{equation}
    \expectedreturn(\policy) =\expectation_{\statezero,\actionzero,\dots}\bigg[\sum\limits_{t=0}^\infty \discountfactor^t \reward(\statet) \bigg], \text{ where } \nonumber
    \statezero \sim \statedistribution_0(\statezero), \actiont \sim \policy(\actiont|\statet), \statetplusone \sim \transitionfunc(\statetplusone|\statet,\actiont). \nonumber
\end{equation}              

where, $\statedistribution_0$ is the state distribution of the initial state $\statezero$. Then, we use the standard definition of the state-action value function $\qfunctionpi$, the state value function $\valuefunctionpi$, advantage function $\advantagepi$, and the unnormalized discount visitation frequencies $\rho_{\pi}$:
\begin{equation}
    \qfunctionpi(\statet,\actiont)=\expectation_{\statetplusone,\actiontplusone,\dots}\bigg[\sum\limits_{l=0}^\infty\discountfactor^l\reward(\state_{t+l})\bigg], \nonumber
\;\;\;\;
    \valuefunctionpi(\statet)=\expectation_{\actiont,\statetplusone,\dots}\bigg[\sum\limits_{l=0}^\infty\discountfactor^l\reward(\state_{t+l})\bigg]. \nonumber
\end{equation}
\begin{equation}
    \advantagepi(\state,\action)=\qfunctionpi(\state,\action)-\valuefunctionpi(\state), \nonumber 
\;\;\;\;
\rho_{\pi}(s) = \sum\limits_{t=0}^{\infty} \gamma^t Pr(s_t = s \vert \pi, \rho_0 ) \nonumber
\end{equation}
where in the definition of $\rho_{\pi}$, $s_0 \sim \rho_0$ and the actions are chosen according to $\pi$.

In~\cite{kakade2002approximately}, the authors derived an expression for the expected return of the another policy $\newpolicy$ in terms of advantage over $\policy$, accumulated over timesteps:
\begin{eqnarray}\label{eqn:approx_return}
    \expectedreturn(\newpolicy)&=\expectedreturn(\policy)+ \expectation_{\statezero,\actionzero,\dots,\sim\newpolicy}\bigg[\sum\limits_{t=0}^\infty \discountfactor^t\advantagepi(\statet,\actiont)\bigg] 
    =\expectedreturn(\policy)+\sum\limits_\state \rho_{\tilde{\pi}} (\state) \sum\limits_\action \newpolicy(\action|\state)\advantagepi(\state,\action). 
\end{eqnarray}
A local approximation to $\expectedreturn(\newpolicy)$ can then be obtained by making an approximation of the state-visitation frequency using the policy $\policy$ which is expressed as
\begin{equation}
    L_\policy(\newpolicy)=\expectedreturn(\policy)+\sum\limits_\state \rho_\policy(\state) \sum\limits_\action \newpolicy(\action|\state)\advantagepi(\state,\action). \nonumber
\end{equation}
In~\cite{schulman2015trust}, the authors present an algorithm to maximize $L_\policy(\newpolicy)$ using a constrained optimization approach. For simplicity, we denote $L_\policy(\newpolicy)$ as $L_{\optparameterold}(\optparameter)$, where $\theta$ represents the policy parameters.
\subsection{Trust Region Policy Optimization (TRPO)}
In this section, we first describe the original TRPO problem and then we present our proposed method to contrast the difference in the optimization techniques. Using several simplifications to the conservative iteration proposed in~\cite{kakade2002approximately}, authors in~\cite{schulman2015trust} proposed a practical algorithm for solving the policy gradient problem using generalized advantage estimation~\cite{schulman2015high}. 
In the TRPO, the following constrained problem is solved at every iteration:
\begin{equation}
  \text{ maximize } L_{\optparameterold}(\optparameter) \text { subject to } \dklestimate (\optparameterold,\optparameter)\leq \delta \nonumber
\end{equation}
where $L_{\optparameterold}(\optparameter)$ is the following term.
\begin{equation}
    L_{\optparameterold}(\optparameter)=\sum\limits_{\state} \rho_{\optparameterold} (\state) 
    \sum\limits_{\action} \policy_{\optparameter} (\action|\state)\advantage_{\pi_{\optparameterold}} 
    (\state,\action) \nonumber
\end{equation}
For simplicity of notation, we will denote $L_{\optparameterold}(\optparameter)$ as $L(\optparameter)$ in the following text. The optimization algorithm in TRPO works in two steps: $(1)$ compute a search direction, using a linear model of the objective and quadratic model to the constraint; and $(2)$ perform a line search in that direction, ensuring that we improve the nonlinear objective while satisfying the nonlinear constraint. The search direction in TRPO and its variants is $\Delta \theta = \alpha F^{-1} g$ where $g = \nabla L(\optparameter)$ is gradient of $L(\optparameter)$ evaluated at $\optparameterold$ and 
$F$ is the Fisher information matrix, i.e., the quadratic model to the KL divergence constraint $\dklestimate (\optparameterold,\optparameter) =\frac{1}{2}(\optparameter-\optparameterold)^T F (\optparameter-\optparameterold)$ and $F$ is the Hessian of the KL divergence estimation evaluated at $\optparameterold$.

\begin{figure}[t]
    \centering
    \includegraphics[width=0.5\linewidth]{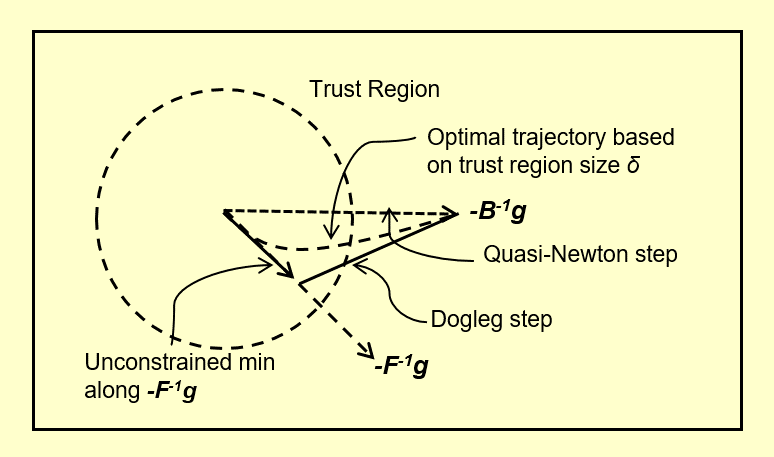}
    \caption{Exact and Dogleg approximation for Trust Region Optimization}
    \label{fig:dogleg_approximation}
\end{figure}

In contrast, the proposed algorithm approximates the objective by a quadratic model and uses the Dogleg method~\cite{NocedalWrightBook} to compute a step.  
Figure~\ref{fig:dogleg_approximation} depicts the idea behind the Dogleg approximation for the trust region optimum. As seen in Figure~\ref{fig:dogleg_approximation} the Dogleg method smoothly transitions between the scaled gradient step and a Quasi-Newton step, which is the unconstrained minimizer of the quadratic model.  
Thus, the step automatically changes direction depending on the size of the trust region.  The size of the trust region is modified according to the accuracy of the quadratic model to ensure global convergence of the algorithm.


\section{Quasi-Newton Trust Region Method (\qntrm)}\label{subsec:TRPM}
\qntrm\, has three distinctive elements that sets it apart from TRPO.  First, the use of a quadratic approximation for the objective via a Quasi-Newton approximation of the Hessian.   Second, the Dogleg method that defines the step. Finally, the adaptive change of the stepsize through the classical trust region framework. We describe each of these in the following.  In the rest of the paper, 
let $f(\optparameter) = -L(\optparameter)$ so that maximization of $L(\optparameter)$ can be equivalently expressed as minimization of $f(\optparameter)$.  We use $\optparameter_k$ to refer to the value of the parameters at the $k$-th iterate of the algorithm. For sake of brevity, $f_k$ denotes $f(\optparameter_k)$, $\nabla f_k$ denotes $\nabla f(\optparameter_k)$ and $\nabla^2 f_k$ denotes $\nabla^2 f(\optparameter_k)$.

\subsection{Quadratic Approximation via BFGS}

\qntrm\, approximates the objective using a quadratic model $f^q_k(\optparameter)$ defined as 
\[
f^q_k(\optparameter) = f_k + \nabla f_k^T (\optparameter - \optparameter_k) + \half (\optparameter - \optparameter_k)^T B_k (\optparameter - \optparameter_k)
\]
where $B_k \approx \nabla^2 f_k$ is  an approximation to the Hessian of $f$ at the point $\optparameter_k$.  We employ the BFGS approximation~\cite{NocedalWrightBook} to obtain $B_k$.  Starting with an initial symmetric positive definite matrix $B_0$, the approximation $B_{k+1}$ for $k \geq 0$ is updated at each iteration of the algorithm using the step $s_k$ and a difference of the gradients of $f$ along the step $y_k = \nabla f(\optparameter_k + s_k) - \nabla f_k$. The update $B_{k+1}$ is the smallest update (in Frobenius norm $\|B - B_k\|_F$) to $B_k$ such that $B_{k+1} s_k = y_k$ (i.e. the secant condition holds), and $B_{k+1}$ is symmetric positive definite, i.e.   
\[
    B_{k+1} = \arg\min\limits_{B} \|B - B_k\|_F \text{ subject to } B s_k = y_k,\; B = B^T.
\]
The above minimization can be solved analytically~\cite{NocedalWrightBook} and the update step is
\begin{equation}
	B_{k+1} = B_{k} - \dfrac{B_k s_k s_k^T B_k}{s_k^T B_k s_k} + \dfrac{y_k y_k^T}{y_k^Ts_k} 
	\label{BFGS_update}
\end{equation}
Observe the effort involved in performing the update is quite minimal. The above update does not enforce positive definiteness of $B_{k+1}$.  By recasting~\eqref{BFGS_update} after some algebraic manipulation as
\[ 
    B_{k+1} = \left(I - \frac{1}{s_k^T B_k s_k} B_ks_ks_k^T\right) B_k \left(I - \frac{1}{s_k^T B_k s_k}s_ks_k^TB_k\right) + \dfrac{y_k y_k^T}{y_k^Ts_k}
\]
it is easy to see that $B_{k+1}$ is positive definite as long as $y_k^Ts_k > 0$. 

\subsection{Dogleg Method}

The search direction in \qntrm\, $\Delta \optparameter_k$ is computed by approximately solving 
\begin{equation}
  \min\limits_{\Delta \optparameter} f^q_k(\optparameter_k+\Delta\optparameter) \text { subject to } (\Delta \optparameter)^T F_k (\Delta \optparameter) \leq \delta_k \nonumber
\end{equation}
i.e. minimizing the quadratic model of the objective subject to the KL-divergence constraint.  The above problem is only solved approximately since the goal is only to produce a search direction $\Delta \optparameter_k$ that furthers the overall objective of minimizing $f(\optparameter)$ at moderate computational cost.  However, the search direction $\Delta \optparameter_k$ should incorporate both the curvature and attain sufficient progress towards a solution. In fact, we desire at least as much progress as the step in TRPO.  The Dogleg method does precisely this by combining the scaled gradient 
direction $\Delta \optparameter^{GD}_k = -\beta_k F_k^{-1} \nabla f_k$ and the QN direction $\Delta \optparameter_k^{QN} = -B_k^{-1} \nabla f_k$.  
The search direction $\Delta \optparameter_k^{DL}$ is obtained using 
Algorithm~\ref{algo:Dogleg}. 

The algorithm first computes the QN direction $\Delta \optparameter_k^{QN}$ and accepts it if the trust region constraint defined by the KL-divergence holds (Step~\ref{Dogleg:qnstep}).  
%
If not the algorithm computes the scaled gradient direction (Step~\ref{Dogleg:scgd}) and a stepsize $\beta_k$ so as to minimize the quadratic model, i.e. 
\begin{equation}
	\beta_k = 
	\frac{\nabla f_k^T F_k^{-1} \nabla f_k}{(F_k^{-1}\nabla f_k)^T B_k (F_k^{-1}\nabla f_k)}. \label{defbetak}
\end{equation}
Unlike the TRPO, observe that due to the curvature in the objective we can now define an \emph{optimal stepsize} for the gradient direction.  If the gradient direction scaled by the optimal stepsize exceeds the trust region then it is further scaled back until the trust region constraint is satisfied and accepted (Step~\ref{Dogleg:gdstep}).  
%
If neither of the above hold then the direction is obtained as a convex combination of the two directions $\Delta \optparameter(\tau_k) := (\Delta \optparameter_k^{GD} + \tau_k (\Delta \optparameter_k^{QN}-\optparameter_k^{GD}))$.  This is the \emph{Dogleg direction}.  The parameter $\tau_k$ is chosen so that the direction $\Delta \optparameter(\tau_k)$ satisfies the trust region constraint as an equality (Step~\ref{Dogleg:dlstep}). The computation of $\tau_k$ requires finding the roots of a quadratic equation which can be obtained easily.

Note that \qntrm\, requires the solution of linear system in order to compute $B_k^{-1} \nabla f_k$ and $F_k^{-1} \nabla f_k$.  Both of these can be accomplished by the Conjugate Gradient (CG) method since $B_k, F_k$ are both positive definite. Thus, the computation \qntrm\, differs from TRPO by an extra CG solve and hence, comparable in computational complexity.

\begin{algorithm}
 \KwData{$\nabla f_k$, $B_k$, $F_k$, $\delta_k$}
 \KwResult{Dogleg direction $\Delta \optparameter_k^{DL}$}
 Compute QN direction $\Delta \optparameter_k^{QN} = -B_k^{-1} \nabla f_k$\;
 \If{$(\Delta \optparameter_k^{QN})^T F_k (\Delta \optparameter_k^{QN}) \leq \delta_k$}{
 	\Return $\Delta \optparameter_k^{QN}$
 }\label{Dogleg:qnstep}
 Compute Gradient direction $\Delta \optparameter_k^{GD} = -\beta_k F_k^{-1} \nabla f_k$ where $\beta_k$ is defined in~\eqref{defbetak}\; \label{Dogleg:scgd}
 \If{$(\Delta \optparameter_k^{GD})^T F_k (\Delta \optparameter_k^{GD}) \geq \delta_k$}{
 	\Return $\sqrt{\frac{\delta_k}{(\Delta \optparameter_k^{GD})^T F_k (\Delta \optparameter_k^{GD})}}\Delta \optparameter_k^{GD}$ \label{Dogleg:gdstep}
 }
 Find largest $\tau_k \in [0,1]$ such that $\Delta \optparameter(\tau_k) := (\Delta \optparameter_k^{GD} + \tau_k (\Delta \optparameter_k^{QN}-\optparameter_k^{GD}))$ satisfies 
 $(\Delta \optparameter(\tau_k))^T F_k (\Delta \optparameter(\tau_k)) = \delta_k$\;
 \Return  $(\Delta \optparameter_k^{GD} + \tau_k (\Delta \optparameter_k^{QN}-\optparameter_k^{GD}))$\;\label{Dogleg:dlstep}
 \caption{Dogleg Method}\label{algo:Dogleg}
\end{algorithm}

\subsection{Trust Region Algorithm}

\qntrm\, combines the curvature information from QN approximation and Dogleg step within the framework of the classical trust region algorithm.  The algorithm is provided in Algorithm~\ref{algo:qntrm} and incorporates safeguards to ensure that $B_k$'s are all positive definite. At each iteration of the algorithm, a step $\Delta \optparameter^{DL}_k$ is computed using Algorithm~\ref{algo:Dogleg} (Step~\ref{qntrm:step}).  The trust region algorithm accepts or rejects the step based on a measure of how well the quadratic model approximates the function $f$ along the step $\Delta \optparameter^{DL}_k$. The commonly used measure~\cite{NocedalWrightBook} is the ratio of the actual decrease in the objective 
and the decrease that is predicted by the quadratic model (Step~\ref{qntrm:ratio}).  If this ratio $\nu_k$ is close to or larger than $1$ then the step computed using the quadratic model provides a decrease in $f$ that is comparable or much better than predicted by the model. The algorithm uses this as an indication that the quadratic model approximates $f$ well. Accordingly, if the ratio (Step~\ref{qntrm:ratio}) is larger than a threshold ($\urho$), the parameters are updated (Step~\ref{qntrm:accept}). If in addition, the ratio is larger than $\orho$ and $\Delta \optparameter_k$ satisfies the trust region size as an equality then the size of the trust region is increased in the next iteration (Step~\ref{qntrm:incdelta}).  This condition indicates that the quadratic model matches the objective $f$ with high accuracy and that the progress is being impeded by the size of the trust region. Hence, the algorithm increases the trust region for the next iteration. With the increased trust region size the algorithm promotes the possible acceptance of a direction other than the scaled gradient direction. On the other hand, if the ratio is below $\urho$ then the computed direction is rejected (Step~\ref{qntrm:reject}) and the size of the trust region is decreased (Step~\ref{qntrm:decdelta}).  This reflects the situation that the quadratic model does not the capture the objective variation well.  Note that as the size of the trust region decreases the performance of the algorithm mirrors that of TRPO very closely. Thus, \qntrm\, is naturally designed to be no worse than the TRPO and often surpass TRPO's performance whenever the quadratic model approximates the objective function well.  Finally, we update the QN approximation whenever the $s_k^Ty_k$ is greater than a minimum threshold.  This ensures that the matrices $B_k$ are all positive definite (Step~\ref{qntrm:update}).  Note that this safeguard is necessary since the Dogleg step cannot be designed to ensure that $s_k^Ty_k > 0$.

\begin{algorithm}
\KwData{Parameters of algorithm -- $0 < \urho < \orho < 1$, $\odelta \in (0,1)$, $\ukappa \in (0,1)$, $0 < \uomega < 1 < \oomega$.} 
\KwData{Initial policy parameters -- $\optparameter_0$}
\KwData{Convergence tolerance -- $\epsilon > 0$, Limit on iterations $K$}
\KwResult{$\optparameter^*$}
Set $k = 0$\;
\While{$\| \nabla f_k \| > \epsilon$ and $k < K$}{
	Compute the Dogleg step $\Delta \optparameter^{DL}_k$ using Algorithm~\ref{algo:Dogleg}\;\label{qntrm:step}
	Compute $\nu_k = \frac{f(\optparameter_k + 	\Delta \optparameter^{DL}_k) - f(\optparameter_k)}{f^q_k(\optparameter_k + 	\Delta \optparameter^{DL}_k) - f^q_k(\optparameter_k)}$\;\label{qntrm:ratio}
	\eIf{$\nu_k \geq \urho$}{
		Set $\optparameter_{k+1} = \optparameter_k + \Delta\optparameter_k^{DL}$\;\label{qntrm:accept}
		\If{$\nu_k \geq \orho$ and $(\Delta \optparameter_k^{DL})^T F_k (\Delta \optparameter_k^{DL}) = \delta_k$}{
			Set $\delta_{k+1} = \min(\odelta,\oomega \cdot \delta_k)$\;\label{qntrm:incdelta}
		}
	}{
		Set $\optparameter_{k+1} = \optparameter_k$\;\label{qntrm:reject}
		Set $\delta_{k+1} = \uomega \cdot \delta_k$\;\label{qntrm:decdelta}
	}
	Set $s_k = \Delta \optparameter_k^{DL}$ and $y_k = \nabla f(\optparameter_k + 	\Delta \optparameter^{DL}_k) - \nabla f(\optparameter_k)$\;
	\eIf{$s_k^Ty_k \geq \ukappa$}{
		Update $B_{k+1}$ using~\eqref{BFGS_update}\;\label{qntrm:update}
	}{
	    Set $B_{k+1} = B_k$\;
	}
	Set $k = k+1$\;
}
\Return $\optparameter^* = \optparameter_k$
\caption{Quasi-Newton Trust Region Method (\qntrm)}\label{algo:qntrm}
\end{algorithm}


\section{Quasi-Newton Trust Region Policy Optimization (\qntrpo)}\label{subsec:TRPO}
\qntrpo\, is the trust region algorithm that we propose in this paper for policy optimization, The algorithm differs from TRPO in the step that is computed at every iteration of policy iteration. For completeness of the paper, it is presented as an Algorithm~\ref{algo:qntrpo}. It is noted that the only difference between \qntrpo\, and TRPO is the way the trust region optimization problem is solved (see line $4$ in Algorithm~\ref{algo:qntrpo}). It is noted that in the original TRPO formulation, the line $4$ in Algorithm~\ref{algo:qntrpo} is performed using the scaled gradient method as discussed earlier. This is the major difference between the proposed and the algorithm proposed in TRPO. Note that QNTRM is an iterative procedure and that the step for every iteration of Algorithm~\ref{algo:qntrpo} is computed by iterating over $K$ steps of QNTRM (see Algorithm~\ref{algo:qntrm}).  This is yet another difference over TRPO where a single gradient descent step is computed for each episode.  As a result, the computational time per episode for QNTRPO is no more than $(2 \times K)$ that of TRPO owing to the possibly two linear systems solves in Dogleg method and K iterations in QNTRM. 

\begin{algorithm}
 Initialize policy parameters ${\optparameter^0}$\\
 \For{$i=0,1,2,\dots$ until convergence}{{Compute all Advantage values $A_{\policy_{\theta^i}}(\state,\action)$ and state-visitation frequency $\rho_{\theta^i}$}\;
 {Define the objective function for the episode $L_{\theta^i}(\theta) = -f^i(\theta)$}\;
 {Obtain  $\optparameter^{i+1}$ using QNTRM to minimize $f^i(\theta)$ with initial policy parameters  $\optparameter_0=\optparameter^i $}}\; 
 \caption{\qntrpo}\label{algo:qntrpo}
\end{algorithm}

\section{Experimental Results}\label{sec:experiments}
In this section, we present experimental results for policy optimization using several different environments for continuous control from the openAI Gym benchmark~\cite{brockman2016openai}. In these experiments, we try to answer the following questions:
\begin{enumerate}
    \item Can \qntrpo\, achieve better learning rate (sample efficiency) than TRPO consistently over a range of tasks?
    \item Can \qntrpo\, achieve better performance than TRPO over a range of tasks in terms of average reward?
\end{enumerate}
In the following, we try to answer these two questions by evaluating our algorithm on several continuous control tasks. In particular, we investigate and present results on four different environments in Mujoco physics simulator~\cite{todorov2012mujoco}. We implement four locomotion tasks of varying dynamics and difficulty: Humanoid~\cite{tassa2012synthesis, duan2016benchmarking}, Half-Cheetah~\cite{heess2015learning}, Walker~\cite{levine2013guided} and Hopper~\cite{schulman2015trust}. The goal for all these tasks is to move forward as quickly as possible. These tasks have been proven to be challenging to learn due to the high degrees of freedom of the robots~\cite{duan2016benchmarking}. A great amount of exploration is needed to learn to move forward without getting stuck at local minima. During the initial learning stages, it is easy for the algorithm to get stuck in a local minima as the controls are penalized and the robots have to avoid falling. The state and action dimensions of these tasks are listed in Table~\ref{tab:task_list}. 

\begin{figure*}
    \begin{multicols}{4}
    \centering
        \subcaptionbox{Humanoid-v2 \label{fig:humanoid_v2}}
            {\includegraphics[width=\linewidth]{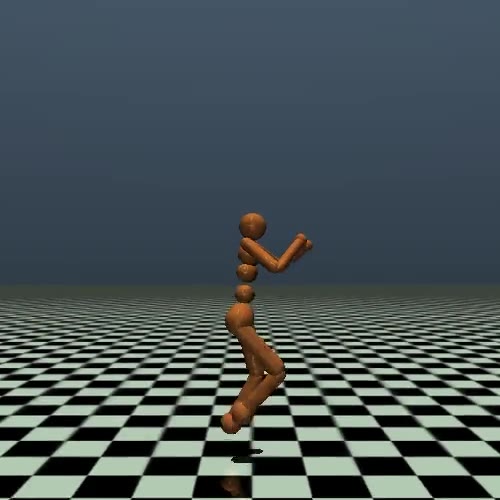}}\par
        \subcaptionbox{HalfCheetah-v2\label{fig:halfcheetah_v2}}
            {\includegraphics[width=\linewidth]{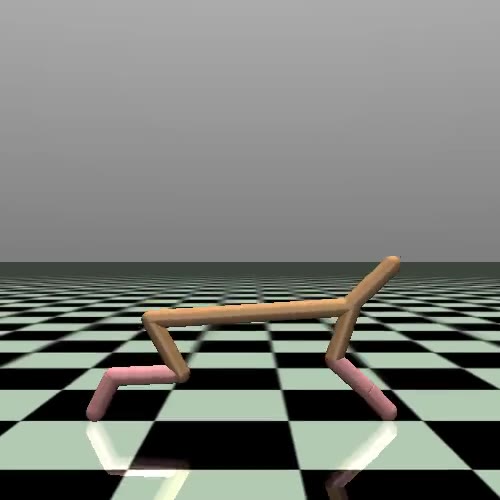}}\par
        \subcaptionbox{Hopper-v2\label{fig:hopper_v2}}
            {\includegraphics[width=\linewidth]{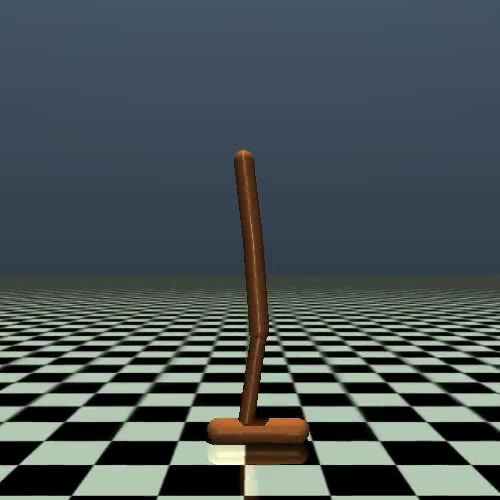}}\par
            \subcaptionbox{Walker 2d-v2\label{fig:walker2d_v2}}
            {\includegraphics[width=\linewidth]{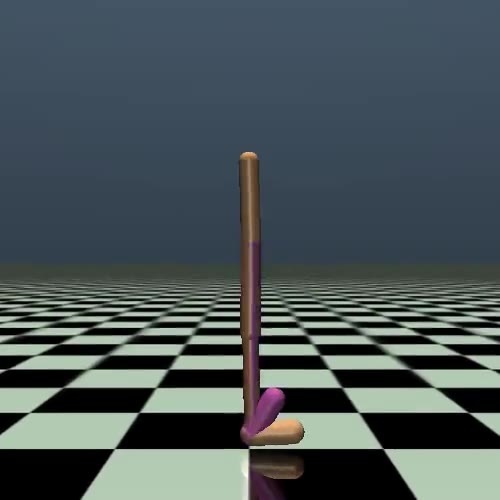}}\par
    \end{multicols}
    \caption{
        The four continuous control benchmark tasks considered in this paper.}
    \label{fig:benchmark_environments}
\end{figure*}


\begin{figure*}
    \begin{multicols}{2}
    \centering
        \subcaptionbox{Humanoid-v2 \label{fig:comparison_humanoid}}
            {\includegraphics[width=1.\linewidth]{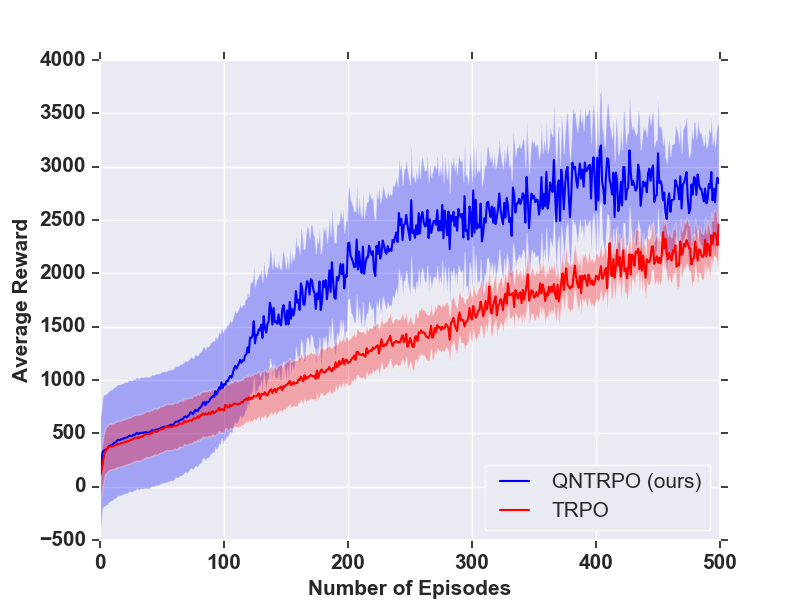}}\par
        \subcaptionbox{HalfCheetah-v2\label{fig:comparison_halfcheetah}}
            {\includegraphics[width=1.\linewidth]{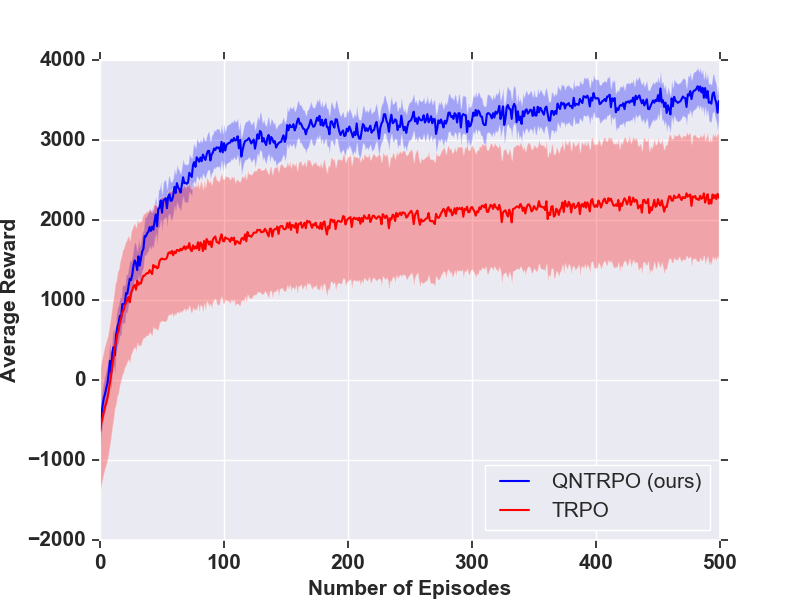}}\par
        \subcaptionbox{Hopper-v2\label{fig:comparison_hopper}}
            {\includegraphics[width=1.\linewidth]{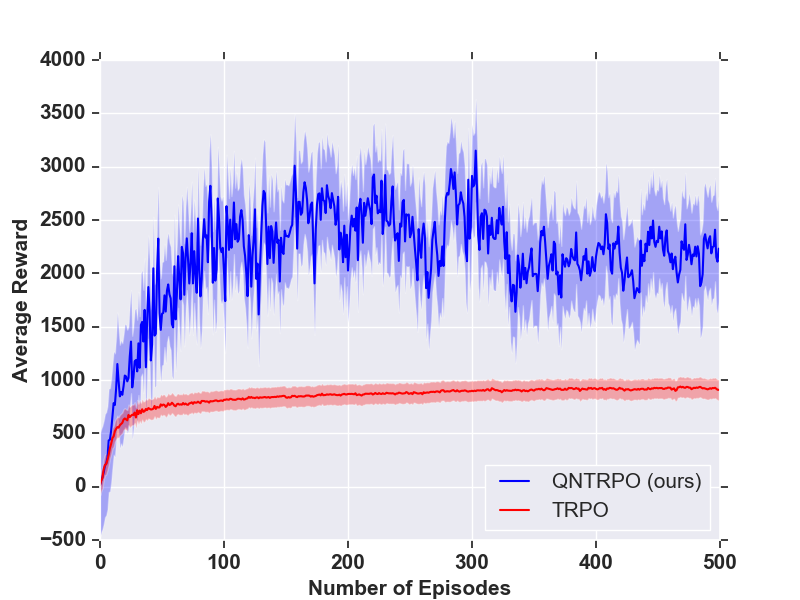}}\par
            \subcaptionbox{Walker 2d-v2\label{fig:comparison_walker2d}}
            {\includegraphics[width=1.\linewidth]{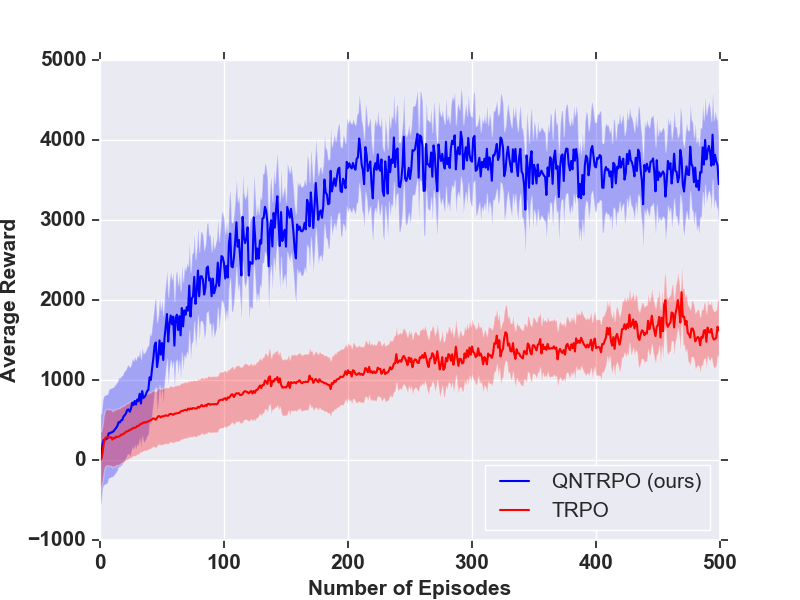}}\par
    \end{multicols}
    \caption{
        Results of our method compared against the TRPO method in~\cite{schulman2015trust} compared on four benchmark continuous control environments in OpenAI gym. The plots show the average batch reward obtained by both methods averaged over three different runs.}
    \label{fig:comparison_with_trpo}
\end{figure*}

\begin{table}[h]
    \centering
    \begin{tabular}{ c | c | c | c | c |  }
     \hline
     \hline
      & Humanoid-v2 & HalfCheetah-v2 & Walker2d-v2 & Hopper-v2  \\ 
     \hline
     State Dimension & 376  & 17 & 17 & 11 \\ 
     Action Dimension & 17 & 6 & 6 & 3 \\ 
    \hline
    \hline
    \end{tabular}
    \caption{State and action dimensions of the RL tasks considered in the paper.}
    \label{tab:task_list}
\end{table}

We run both TRPO and \qntrpo\, for 500 episodes and average all results across five different runs with different random seeds for the environment initialization. All hyperparameters for the algorithms -- batch size, policy network architecture, step size and the generalized advantage estimation coefficient ($\lambda$) -- are identical for both algorithms. As TRPO (and thus \qntrpo\,) performs better with bigger batches, we use a batch size of $15000$. In each of these episodes, trajectories are generated for a maximum length of $2000$ and then restarted either if the terminal condition is met or the trajectory length is satisfied. The network architecture is kept the same across all the tasks. The trust region radius is chosen to be $0.1$ (note that this is the parameter $\bar{\delta}$ in Algorithm~\ref{algo:qntrm}). At lower trust region radius both algorithms performed slower and thus the results are not reported here. The discount factor $\discountfactor$ is chosen to be $0.99$ and the constant $\lambda$ for advantage function estimation is chosen to be $0.97$. The parameters for \qntrm\, were chosen to be the following: 
$K = 10$, $\orho = 0.75$, $\urho = 0.1$, $\uomega = 0.3$, $\oomega = 2$ and $\ukappa = 10^{-3}$. 
Codes for running these experiments are available at \href{www.merl.com/research/license#QNTRPO}{www.merl.com/research/license\#QNTRPO}.

Results of our experiments are shown in Figure~\ref{fig:comparison_with_trpo}. For all four tasks, we can demonstrate that \qntrpo\, can achieve faster learning, and thus better sample efficiency than the original TRPO. Furthermore, the performance of \qntrpo\, is also significantly better than TRPO. This is evident from the fact that \qntrpo\, achieves higher rewards than TRPO, which also has longer transitory. For high complexity problems like Humanoid, QNTRPO takes about $350$ episodes with the current batch size to reach the maximum score (of around $3000$). These results show that \qntrpo\, can calculate a better step for the constrained optimization problem for policy iteration using \qntrm. 

\section{Conclusions and Future Work}\label{sec:conclusions}
In this paper, we presented an algorithm for policy iteration using a Quasi-Newton trust region method. The problem was inspired by the policy optimization problem formulated in~\cite{schulman2015trust} where a linesearch is performed to compute the step size in the direction of steepest descent using a quadratic model of the constraint. In this paper, we proposed a dogleg method for computing the step during policy iteration which has theoretical guarantees~\cite{NocedalWrightBook} of better performance over the scaled gradient descent method used in~\cite{schulman2015trust}. The proposed method was compared against the original TRPO algorithm in four different continuous control tasks in Mujoco physics simulator. The proposed algorithm outperformed TRPO in learning speed as well performance indicating that the proposed method can compute better step for the policy optimization problem. 

Despite the good performance , there are a number of open issues for which we do not have a complete understanding. We have observed that the maximum trust region radius ($\odelta$) plays an important role in speed of learning.  However, choosing this  arbitrarily might result in poor convergence. Furthermore, to achieve monotonic improvement in policy performance, one has to select the trust region radius very carefully which is undesirable. 
It would also be interesting to study 
the interplay of batch size and trust region radius. This can help address the issue of steplength selection. In the future, we would like to further investigate several features of the proposed algorithm including the following.
\begin{itemize}
    \item Analyze the stability of the proposed algorithm to the size of trust region radius and batch size. We believe that the proposed method can be used to fine tune the hyperparameter of trust region radius which controls the maximum step size in each iteration of the algorithm.
    \item Evaluate the proposed algorithm on much higher dimension learning problem for end-to-end learning using a limited memory version of the proposed algorithm.
    \item Use ideas from ensemble methods~\cite{rajeswaran2016epopt}, scalable bootstrapping~\cite{kleiner2014scalable} and factored methods to curvature~\cite{wu2017scalable} for better and efficient approximation of the objective function.
    \item Evaluate the performance on challenging, sparse reward environments~\cite{baar2019, romeres2019}.
\end{itemize}


\bibliography{main}  

\appendix{}
\section{Derivation of the Dogleg step for \qntrm}
The Dogleg method aims to obtain an approximate solution of the trust region problem
\begin{equation}
  \min\limits_{\Delta \optparameter} f^q_k(\optparameter_k+\Delta\optparameter) \text { subject to } (\Delta \optparameter)^T F_k (\Delta \optparameter) \leq \delta_k \label{stepopt}
\end{equation}
where $f^q_k(\optparameter_k+\Delta\optparameter) = f_k + \nabla f_k^T (\Delta \optparameter) + \half (\Delta \optparameter)^T B_k (\Delta \optparameter)$. 
In this section, we derive the Dogleg step under the trust region defined by the KL-divergence constraint.  

We begin by first transforming the trust region problem in~\eqref{stepopt} into standard form.  Let $F_k = L_k L_k^T$ which can be obtained for example by Cholesky factorization since the Fischer matrix $F_k$ is positive definite.  Note that the factorization is only used for deriving the step and is never required for the computations.

Defining $\widehat{\Delta \optparameter} = L_k^T \Delta \optparameter$ we can recast the quadratic model as
\begin{equation}
    \widehat{f}_k^q(\optparameter_k + \widehat{\delta \optparameter}) = f_k + \widehat{\nabla f}_k^T (\widehat{\Delta \optparameter}) + \half (\widehat{\Delta \optparameter})^T \widehat{B}_k (\widehat{\Delta \optparameter})
\end{equation}
where $\widehat{\nabla f}_k = L_k^{-1} \nabla f_k$ and 
$\widehat{B}_k = L_k^{-1}B_kL_k^{-T}$. It is easily verified that $f^q_k(\optparameter_k+\Delta\optparameter) = \widehat{f}^q_k(\optparameter_k+L_k^T\Delta\optparameter)$ and 
$(\Delta \optparameter)^T F_k (\Delta \optparameter) = (\widehat{\Delta \optparameter})^T (\widehat{\Delta \optparameter})$. Hence, the trust region problem in~\eqref{stepopt} can be recast as the standard trust region problem
\begin{equation}
  \min\limits_{\widehat{\Delta \optparameter}} \widehat{f}^q_k(\optparameter_k+\widehat{\Delta\optparameter}) \text { subject to } (\widehat{\Delta \optparameter})^T  (\widehat{\Delta \optparameter}) \leq \delta_k \label{eqstepopt}
\end{equation}
In the following, we will derive the Quasi-Newton, Gradient and Dogleg steps based on~\eqref{eqstepopt} and then, transform these steps to the original space using the transformation $\widehat{\Delta \optparameter} = L_k^T \Delta \optparameter$.

The Quasi-Newton step for~\eqref{eqstepopt} is 
\begin{equation}
    \widehat{\Delta \optparameter}^{QN} = -\widehat{B}_k^{-1}\widehat{\nabla f}_k = - L_k^T B_k^{-1}\nabla f_k \label{eqQNstep}
\end{equation}
where the second equality is obtained by substitution.  Thus, the Quasi-Newton step in the original space of parameters is 
\begin{equation}
    \Delta \optparameter^{QN} = - B_k^{-1} \nabla f_k.
\end{equation}

The gradient direction for~\eqref{eqstepopt} is $\widehat{\Delta \optparameter}^{gd} = - \widehat{\nabla f}_k = -L_k^{-1}\nabla f_k$. The optimum stepsize $\beta_k$ along the gradient direction is obtained from 
\begin{equation}
    \min\limits_{\beta} \widehat{f}_k^q(\optparameter_k + \beta \widehat{\Delta \optparameter}^{gd}).
\end{equation}
Hence, the optimal stepsize along the gradient direction is 
\begin{equation}
    \beta_k = \frac{\widehat{\nabla f}_k^T\widehat{\nabla f}_k}{\widehat{\nabla f}_k^T\widehat{B}_k \widehat{\nabla f}_k} = \frac{\nabla f_k^T F_k^{-1}\nabla f_k}{(F_k^{-1}\nabla f_k)^T B_k (F_k^{-1}\nabla f_k)}
\end{equation}
and the scaled gradient direction is
\begin{equation}
    \widehat{\Delta \optparameter}^{GD} = -\beta_k \widehat{\nabla f}_k.
\end{equation}
Thus, the scaled gradient step in the original space of parameters is
\begin{equation}
    \Delta \optparameter^{GD} = -\beta_k F_k^{-1} \nabla f_k.
\end{equation}

The Dogleg step for~\eqref{eqstepopt} computes a $\tau_k$ such that 
\begin{equation}
\begin{aligned}
    & \left\| \widehat{\Delta \optparameter}^{GD} + \tau_k 
    (\widehat{\Delta \optparameter}^{QN} - \widehat{\Delta \optparameter}^{GD}) \right\|^2 &&= \delta_k \\
\implies & \left\| L_k^T \Delta \optparameter^{GD} + \tau_k 
    (L_k^T\Delta \optparameter^{QN} - L_k^T \Delta \optparameter^{GD}) \right\|^2 &&= \delta_k \\
\implies & \Delta \optparameter(\tau_k) F_k \Delta \optparameter(\tau_k) &&= \delta_k     
\end{aligned}
\end{equation}
where $\Delta \optparameter(\tau_k) = \Delta \optparameter^{GD} + \tau_k (\Delta \optparameter^{QN} - \Delta \optparameter^{GD})$.

\section{Time performance comparison}

In Table~\ref{tab:task_clocktime} we compare the wall clock time for each of the four tasks. For each task we average the time needed to perform each single episode over all the episodes. The performance are computed on a Linux desktop with i7-6700K Intel Core.

\begin{table}[h]
    \centering
    \begin{tabular}{ | c | c | c | c | c |  }
     \hline
     \hline
     Algorithm & Humanoid-v2 & HalfCheetah-v2 & Hopper-v2 & Walker2d-v2 \\ 
     \hline
     TRPO &  9.68 $\pm$  0.13\, [s] & 3.19 $\pm$ 0.018  \, [s] & 3.79 $\pm$ 0.04  \, [s] & 4.29 $\pm$ 0.06  \, [s] \\ 
     QNTRPO & 91.99 $\pm$ 10.87 \, [s] & 54.66 $\pm$ 8.65  \, [s] & 30.02 $\pm$ 7.22  \, [s] & 37.98 $\pm$ 6.78  \, [s] \\ 
    \hline
    \hline
    \end{tabular}
    \caption{Average and standard deviation in seconds of wall clock time for each episode of all the experiments for the 4 environments on a Linux desktop with i7-6700K Intel Core.}
    \label{tab:task_clocktime}
\end{table}

QNTRPO is slower than the standard TRPO due to multiple inner iteratins that are performed for each episode. The time performance is consistent with the computational analysis described in the paper.

The QNTRM is an iterative procedure and the step for every iteration of Algorithm~3 is computed by iterating over $K$ steps of QNTRM (see Algorithm~2).  Instead, in TRPO a single gradient descent step is computed for each episode.  As a result, the computational time per episode for QNTRPO is no more than $(2 \times K)$ that of TRPO owing to the possibly two linear systems solves in Dogleg method and K iterations in QNTRM. In our experiments K is chosen to be 10 and it is clear from Table~\ref{tab:task_clocktime} that the ratio in performance time between QNTRPO and TRPO is below 20.


\end{document}